\documentclass[suppldata]{interact}
\usepackage[]{graphicx}
\usepackage[]{color}
\usepackage{algorithm,algpseudocode,float}
\usepackage{subcaption}
\makeatletter
\def\maxwidth{ %
	\ifdim\Gin@nat@width>\linewidth
	\linewidth
	\else
	\Gin@nat@width
	\fi
}
\makeatother

\definecolor{fgcolor}{rgb}{0.345, 0.345, 0.345}

\usepackage{framed}
\makeatletter
{\par\unskip\endMakeFramed%
	\at@end@of@kframe}
\makeatother

\definecolor{shadecolor}{rgb}{.97, .97, .97}
\definecolor{messagecolor}{rgb}{0, 0, 0}
\definecolor{warningcolor}{rgb}{1, 0, 1}
\definecolor{errorcolor}{rgb}{1, 0, 0}

\usepackage{alltt}
\usepackage{amsmath,amsthm,mathtools}
\usepackage{graphics}
\usepackage{algorithm}
\usepackage{algpseudocode}
\usepackage{color,soul}
\usepackage{float}
\usepackage{multicol}
\usepackage[makeroom]{cancel}
\usepackage{comment}
\usepackage[usenames,dvipsnames]{xcolor}
\usepackage[justification=justified,skip=2pt ]{caption}
\DeclareMathOperator*{\argmin}{arg\,min}
\usepackage{epstopdf}
\usepackage{natbib}

\renewcommand\bibfont{\fontsize{10}{12}\selectfont}
\makeatletter
\def\NAT@def@citea{\def\@citea{\NAT@separator}}
\makeatother
\theoremstyle{plain}
\newtheorem{theorem}{Theorem}[section]
\newtheorem{lemma}[theorem]{Lemma}

\theoremstyle{definition}

\theoremstyle{remark}

\IfFileExists{upquote.sty}{\usepackage{upquote}}{}

\begin{document}
	
	\title{Local Polynomial $L_p$-norm Regression}
	
	\author{
		\name{Ladan Tazik \textsuperscript{a}, James Stafford \textsuperscript{b}, John Braun\textsuperscript{c}, }
		\affil{\textsuperscript{a,c} Dept. of Computer Science, Mathematics, Physics and Statistics, University of British Columbia, Kelowna, BC, Canada;\\ \textsuperscript{b}Dept. of Statistical Sciences, University of Toronto, Toronto, ON, Canada}
	}
	\maketitle
	
\begin{abstract}
	
	The local least squares estimator for a regression curve cannot provide optimal results when non-Gaussian noise is present. Both theoretical and empirical evidence suggests that residuals often exhibit distributional properties different from those of a normal distribution, making it worthwhile to consider estimation based on other norms. It is suggested that $L_p$-norm estimators be used to minimize the residuals when these exhibit non-normal kurtosis. In this paper, we propose a local polynomial $L_p$-norm regression that replaces weighted least squares estimation with weighted $L_p$-norm estimation for fitting the polynomial locally. We also introduce a new method for estimating the parameter $p$ from the residuals, enhancing the adaptability of the approach. Through numerical and theoretical investigation, we demonstrate our method's superiority over local least squares in one-dimensional data and show promising outcomes for higher dimensions, specifically in 2D.

\end{abstract}


\section{Introduction}

Given $n$ independent data points $(x_i, y_i)$, we aim to represent the relationship between a predictor variable $x$ and response variable $y$ as
\begin{equation} \label{eq:0}
	y_i = m(x_i)+ \varepsilon_i	\hspace{0.5cm} i = 1,2,..., n.
\end{equation}
Where $m(x)$ is the expected value of $y$ given $x$ and $\varepsilon$ is the error.  The fitted model, denoted as $\hat m(x)$ predicts the response variable $\hat{y}$ as
\begin{equation*}\label{eq:yhat}
	\hat{y_i} =\hat{m}(x_i) \hspace{0.5cm} i = 1,2,..., n.
\end{equation*}

To estimate $m(x)$, Least Squares (LS) method is commonly used, minimizing the sum of squared residuals:
\begin{equation} \label{eq:3}
	\sum_{i=1} ^n \left|y_i - \hat{m}(x_i)\right|^2 = \sum_{i=1}^n  \left|e_i\right|^2.
\end{equation}
LS is essentially equivalent to maximum likelihood estimation (MLE) when the errors are normally distributed, applicable to both parametric and non-parametric models for $m(x)$, offering optimal properties such as the Best Linear Unbiased Estimator (BLUE) under these and additional assumptions \citep{Lovett}. However, real-world data often violate normality assumption and are prone to outliers, which can render LS estimates suboptimal under a Mean Squared Error (MSE) criterion. Thus, exploring nonlinear estimators might offer improvements \citep{anderson1982robust}.

\citet{subbotin} introduced a flexible alternative to the normal distribution, known as the Generalized Error Distribution (GED), which is denoted and formulated in this study as follows \citep{giacalone2020combined}
\begin{equation*}\label{eq:epf}
	f_p (z) = \frac{1}{2p^\frac{1}{p} \sigma_p \Gamma(1+\frac{1}{p})}\; \exp{\left[ -\frac{1}{p}\left|\frac{z-M_p}{\sigma_p}\right|^p\right]},
\end{equation*}
where $M_p = E[z]$, is the location parameter and $\sigma_p = E\left[\left|z-M_p\right|\right]^\frac{1}{p}$, is the scale parameter, and $p>0$ is the shape parameter. Special cases include the Laplace distribution $(p = 1)$ and the Gaussian distribution $(p = 2)$. \par

The $L_p$-norm regression estimator, defined by minimizing the sum of the $p$-th powers of absolute residuals, extends the LS method. $L_p$-norm estimators exhibit lower asymptotic variance and align with MLE when the error distribution follows the GED \citep{gonin1985nonlinear}. Moreover, the $L_p$-norm method has been shown to be efficient in scenarios with diverse multicollinearity degrees \citep{giacalone2018multicollinearity,giacalone2021optimal}. \par 
The selection of $p$ impacts the statistical efficiency of the estimator, influenced by the underlying error distribution characteristics \citep{ricenorms, money1982linear}. Various methods have been introduced for estimating the optimal $p$. These include methods based on sample residual kurtosis ($\kappa$), such as those proposed by \citet{harter1974method}, and empirical formulas suggested by \citet{forsythe1972robust} and \citet{sposito1982unbiased}. However, some methods, like the one proposed by \citet{money1982linear}, as confirmed in our simulations (figure \ref{fig:comp_p}) systematically underestimates $p$ due to the stabilization of $\kappa$ around 2 for light-tailed distributions. To address these limitations \citet{mineo1989norm, mineo1994nuovo} extended Money et al.'s work by using a generalized kurtosis function derived from the GED. Mineo’s approach improved stability across different tail behaviors. However, it still relied on moment-based calculations, which can be sensitive to small sample sizes and extreme values. An alternative approach by \citep{giacalone2020combined}  combined multiple kurtosis-based indexes to refine the estimation of $p$, particularly in forecasting applications. His method improves upon previous moment-based estimators by incorporating additional shape descriptors, making it more robust for certain data distributions.\par 
Generally, local regression provides a flexible alternative to parametric regression  for sufficiently smooth regression functions \citep{fan2018local}. 
The solution of 
\begin{equation}\label{eq:full}
	\min_{\beta} \sum_{i = 1}^n|y_i - \beta_0-\beta_1(x_i -x)-\ldots-\beta_N(x_i-x)^N|^2 K_h(x_i-x) 
\end{equation}
yields an estimator for $\beta_0$, which approximates the regression function, and $\beta_1$, which approximates the first derivative $m'(x)$ at $x_i$. The local least squares (LLS) objective described above involves a kernel function $K_h$ and bandwidth $h$ to weight data points based on their distance from the evaluation point $x$. \par 
Local regression techniques, including the Nadaraya-Watson estimator and LOESS method, have been refined for handling multivariate data and optimizing bandwidth selection to minimize Mean Integrated Squared Error (MISE), with significant contributions from researchers \citep{nadaraya1964estimating, watson1964smooth, gasser1984estimating, stone1977consistent, cleveland1979robust, cleveland1988regression, buja1989linear, fan1992variable, ruppert1994multivariate, ruppert1995effective, xia2002asymptotic, li2004cross}. These advancements facilitate more accurate and robust curve estimation in the presence of complex data structures and boundary effects, contributing to the ongoing development of nonparametric regression methodologies \citep{spokoiny1998estimation,seifert2000data, li2003multivariate, kai2010local, cheng2018bias}.

The remainder of this paper is organized as follows: In Section 2, the methodology for local $L_p$-norm regression is explored and the conceptual foundations of the method are explained. Section \ref{theory}  describes the method's theoretical characteristics.  Our approaches to selecting the bandwidth and the shape parameter $p$ are discussed in Section \ref{sec:bw} and \ref{sec:p}. To illustrate the efficacy of our approach, Section \ref{sec:sim} includes simulated experiments and several real-world applications for one-dimensional data.  The conclusion in Section \ref{sec:conc}  also includes a short discussion of a preliminary simulation study for two-dimensional (2D) data, which shows promising results that underscore the potential for extending this method to higher dimensions.  
\section{Local Polynomial $L_p$Norm Regression}

Consider the LLS objective provided in equation (\ref{eq:full}), in view of the remarks made in the introduction regarding possible MSE-suboptimality of LS.  We propose to investigate what happens when we replace the $L_2$-norm with a specified $L_p$-norm.   Later, we shall consider data-driven methods for selecting $p$ but first, we set out the new local constant and local linear objectives.\par  
We designate the objective functions (\ref{eq:lc}) and (\ref{eq:llp}) as local constant $L_p$-norm regression and local linear $L_p$-norm regression, respectively. In both instances, the local $L_p$-norm estimator of the unknown function $m(.)$ at $x$ is represented as $\beta_0$. Additionally, the  derivative of the function, ${m'}(x)$, is in agreement with the value of $\beta_1$ in function displayed in (\ref{eq:llp}).
\begin{align} 
	&\sum_{i = 1}^{n}|y_i -\beta_0|^p K_h(x_i - x) \label{eq:lc} \\
	&\sum_{i = 1}^{n} |y_i -\beta_0 -\beta_1(x_i - x)|^p K_h(x_i - x) \label{eq:llp}
\end{align}
where $K_h$ denotes a weight function and $h$ is a bandwidth parameter.
Our approach is  equivalent to local likelihood estimation when the errors in the data conform to the GED. A proof is furnished for local linear regression utilizing $L_p$-norm, and the process for local constant $L_p$-norm regression aligns similarly. The likelihood function is expressed as follows:
\begin{small}
\begin{gather*}
	\prod_{i=1}^{n} P(y_i \vert x_i ; \beta_0, \beta_1 , \sigma_p) =\
	\prod_{i=1}^{n} \frac{1}{2p^\frac{1}{p} \sigma_p \Gamma(1+\frac{1}{p})} \exp\left[ -\frac{1}{p}\left|\frac{y_i -\beta_0 -\beta_1(x_i - x)}{\sigma_p}\right|^p\right]
\end{gather*}
\end{small}
To confine our consideration solely to the local data points around $x_i$, we incorporate the weight function $K_h(x_i - x)$ within the density function. Consequently, taking the logarithm of the aforementioned expression yields the local log-likelihood function:
\begin{small}
\begin{gather*}
	\ell(\beta_0, \beta_1) = \sum_{i=1} ^n \log \left( \frac{1}{2p^\frac{1}{p} \sigma_p \Gamma(1+\frac{1}{p})} \exp\left[ -\frac{1}{p}\left|\frac{y_i -\beta_0 -\beta_1(x_i - x)}{\sigma_p}\right|^p K_h(x_i - x)\right] \right) =\\
	-n\log\Big(2p^\frac{1}{p}\sigma_p \Gamma(1+\frac{1}{p})\Big) - \left(\frac{1}{p\sigma_p^p}\right)\sum_{i=1}^{n}\left|y_i -\beta_0 -\beta_1(x_i - x) \right|^p K_h(x_i - x).
\end{gather*}
\end{small}
Maximizing $\ell(\beta_0, \beta_1)$ with respect to $\beta$ yields the local linear $L_p$ norm estimator:

\begin{gather}
	\argmin_{\beta}\Big(\sum_{i=1} ^n\left|y_i -\beta_0 -\beta_1(x_i - x)\right|^p K_h(x_i - x)\Big).
	\label{eq:llp_obj}
\end{gather}

Minimizing the objective function (\ref{eq:llp_obj}) necessitates the use of a nonlinear equation solver. It is worth noting that local quadratic and cubic $L_p$-norm estimators can be derived by increasing the order of the Taylor Expansion to $N = 2$ and $3$, respectively. However, such an extension can lead to computational issues, especially when the design points are sparse in certain regions. Our methodology is outlined in Algorithm \ref{alg:llp} in Appendix A.\par 

To initiate the method, we must first select two tuning parameters: the bandwidth $h$ and the shape parameter $p$.   These issues are addressed in Section \ref{sec:bw}
and Section \ref{sec:p}.

\section{Theoretical Properties} \label{theory}

{In this section, we show that the leading terms of the asymptotic bias for the local constant $L_p$-norm and local linear regression estimators are the same as for the local constant and linear $L_2$-norm estimators, but the asymptotic variances are adjusted according to the moment properties of the error distribution.  }

Following the arguments given by \citet{Nyquist80}, we can also show that the distribution of the new estimator is
asymptotically normal.  

We obtain the asymptotic bias, variance and normality in two steps. These are summarized in the lemma and theorem below.  First, we show that, under sufficiently strong
moment assumptions, a central limit theorem holds for sums of appropriately transformed responses $y_i$.  In the 
second step, convexity of the $L_p$-norm is used, together with the result of the first step, to obtain the 
central limit theorem for the both local $L_p$ norm estimator. The results in the  Theorem \ref{theorem} will be used based on the following assumptions: \\

\begin{enumerate}
\item[A.1] The model for the data is given by equation (\ref{eq:0}). 
\item[A.2] $\varepsilon_1, \varepsilon_2, \ldots, \varepsilon_n$ are independent and identically distributed random
variables with common distribution function $F(\varepsilon)$, $E[|\varepsilon_i|^{p-1} \mbox{sign}(\varepsilon_i)] = 0$, and $E[|\varepsilon_i|^{(p-1)(2+\delta)}] < \infty$, for $i = 1, 2, \ldots, n$. 
\item[A.3] The design points $x_1, x_2, \ldots, x_n$ are
independent random variables distributed according to the probability density function $f(x)$ which has a bounded first derivative which has support on the interval $(a,b)$.
\item[A.4] $K_h$ is a symmetric, nonnegative kernel with scale parameter $h$ and support on $[-h, h]$.  
\item[A.5] $h$ is a function of $n$ with the property that $h \rightarrow 0$ and 
$nh \rightarrow \infty$, as $n \rightarrow \infty$ and $nh^5 = o(1)$.
\end{enumerate}

 We denote $\int x^2 K_1(x) dx$ by $\mu_2(K)$ and $\int K_1^2(x) dx$ by $R(K)$.   

\begin{lemma}\label{lem:clt}  Assume A.2 through A.5.  For any $x \in (a+h, b-h)$, and for $j=0,1$,  define  
	\[V_{jn} = \frac{1}{h^j}\sqrt{\frac{h}{n}}\sum_{i=1}^n (x_i-x)^j|\varepsilon_i|^{p-1}\mbox{sign}(\varepsilon_i)K_h(x_i-x).\]
	Then 
	\begin{equation}\label{eqn:bivlimit} (V_{0n}, V_{1n}) \xrightarrow{\mathcal{D}} N\left( (0,0), \begin{bmatrix}
		R(K) & 0\\ 0& \mu_2(K^2)
	\end{bmatrix}f(x) E[|\varepsilon_1|^{2p-2}]\right)\end{equation}
\end{lemma}

\begin{proof}
	Define, for any real-valued $\alpha_0$ and $\alpha_1$,
	\[S_{nh} = \alpha_0V_{0n}+\alpha_1V_{1n} = \frac{1}{\sqrt{n}}\sum_{i=1}^n \left(\alpha_0\sqrt{h}+\frac{\alpha_1(x_i-x)}{\sqrt{h}}\right)|\varepsilon_i|^{p-1}\mbox{sign}(\varepsilon_i)K_h(x_i-x).\]
	Because of A.2, we have 
	\[E[S_{nh}] = 0\]
	and 
	\begin{small}
	\begin{eqnarray}
		\operatorname{Var}(S_{nh})&= & \int\left(\alpha_0\sqrt{h}+\frac{\alpha_1(z-x)}{\sqrt{h}}\right)^2 E[|\varepsilon_1|^{2p-2}]K_h^2(z-x)f(z)dz \nonumber  \\
		 &= &\int(\alpha_0+\alpha_1v)^2 K^2(v) f(x+vh) dv E[|\varepsilon_1|^{2p-2}] \nonumber  \\
		 &= &(\alpha_0^2R(k)+\alpha_1^2\mu_2(K^2))f(x)E[|\varepsilon_1|^{2p-2}] + O(h). \label{eqn:CW}
	\end{eqnarray}
	\end{small}
	
	It is straightforward to verify the Lyapunov condition, separately for $V_{0n}$ and $V_{1n}$, and
	hence for $S_{nh}$.   Therefore,  $S_{nh}/\sqrt{\operatorname{Var}(S_{nh})}$ converges in distribution to a standard normal random variable. By the Cramér-Wold device, $V_{0n}$ and $V_{1n}$ jointly converge to a bivariate normal distribution.   The mean vector is clearly $(0,  0)$, and by taking $\alpha_0 = 1 - \alpha_1 = 0$, and $\alpha_0 = 1 - \alpha_1 = 1$, respectively, in (\ref{eqn:CW}), we obtain
	the diagonal elements of the asymptotic covariance matrix in (\ref{eqn:bivlimit}).  The Cramér-Wold device
	and the form of (\ref{eqn:CW}) imply that the off-diagonal elements of the asymptotic covariance
	matrix must be 0.  	
\end{proof}


	\begin{theorem} \label{theorem} Suppose assumptions A.1 through A.5 hold, and $p > 1$.  
	\begin{enumerate}
	\item[(i)] If $m(x)$ is a function with two continuous derivatives on $(a,b)$ and $\hat{m}(x)$ is the local constant $L_p$-norm estimator for $m(x)$, then
$\hat{m}(x)$ is asymptotically normal with mean \[m(x) +\mu_2(K)h^2\left(\frac{m'(x)f'(x)}{f(x)} + \frac{m''(x)}{2}\right) + o(h^2), \quad \forall x \in (a+h, b-h)\] and variance \[\frac{R(K)E[|\varepsilon_1|^{2p-2}]}{nh(p-1)^2 E^2[|\varepsilon_1|^{p-2}] f(x)}+  o\left(\frac{1}{nh}\right), \quad \forall x \in (a+h, b-h).\]

\item[(ii)] If $m(x)$ is a function with three continuous derivatives on $(a,b)$ and 
\begin{enumerate}
\item[(a)] if
$\hat{m}(x)$ is the local linear $L_p$-norm estimator for $m(x)$, then 
$\hat{m}(x)$ is asymptotically normal with mean  
\[m(x) + \frac{\mu_2(K)h^2m''(x)}{2} + o(h^2), \quad \forall x \in (a+h, b-h)\]
and variance 
\[\frac{R(K)E[|\varepsilon|^{2p-2}]}{nhf(x)(p-1)^2 E^2[|\varepsilon|^{p-2}]} + o\left(\frac{1}{nh}\right), \quad \forall x \in (a+h, b-h).\]
\item[(b)] if
$\hat{m'}(x)$ is the local linear $L_p$-norm estimator for $m'(x)$, then
$\hat{m'}(x)$ is asymptotically normal with mean 
\[m'(x)+\frac{\mu_4(K)h^2 m'''(x)}{3!\mu_2(K)} + o(h^2), \quad \forall x \in (a+h, b-h).\]
and variance \[\frac{\mu_2^2(K^2)E[|\varepsilon|^{2p-2}]}{nh^3f(x)\mu_2(K)(p-1)^2E^2[|\varepsilon|^{p-2}]} + o\left(\frac{1}{nh^3}\right), \quad \forall x \in (a+h, b-h).\]

\end{enumerate}
\item[(iii)] $\hat{m}(x)$ and $\hat{m'}(x)$ are asymptotically independent.  
\end{enumerate}
\end{theorem}

\begin{proof}
\begin{enumerate}
\item[(i)]
The local constant $L_p$-norm estimator for $\beta_0 = m(x)$ is the quantity $\widehat{\beta}_0$ which minimizes the expression 
	\[ \ell(\beta_0) = \sum_{i=1}^n |y_i - \beta_0|^p K_h(x_i - x). \]  
	Differentiating with respect to $\beta_0$ once, we have
	\begin{equation*}
	\ell'(\beta_0) = -p \sum_{i=1}^n |y_i - \beta_0|^{p-1} \mbox{sign}(y_i - \beta_0) K_h(x_i - x). \hspace{1cm} (*)
	\end{equation*}	
	Observe that since $\ell(.)$ is a convex function, it has a unique minimizer.  Furthermore, $\ell'(\beta_0) < 0$ if,  and only if $\beta_0< \widehat{\beta}_0$, and $\ell'(\beta_0) > 0$ if and only if $\beta_0 > \widehat{\beta}_0$.  Let $F_n(x)$ denote the distribution function of $\sqrt{nh}(\widehat{\beta}_0 - \beta_0)$.  Then
	\begin{equation*}
	 F_n(c) = P(\sqrt{nh} (\widehat{\beta}_0 - \beta_0) \leq c) = P(\widehat{\beta}_0 \leq \beta_0 + c/\sqrt{nh}) = P(\ell'(\beta_0 + c/\sqrt{nh}) \geq 0). \hspace{1cm} (**)
	\end{equation*}	
	
	Substituting (*) for (**), where $\beta_0$ replaced by $\beta_0+ c/\sqrt{nh}$, gives
\begin{small}
\begin{eqnarray} 
	F_n(c) &= & P\left(-p \sum_{i=1}^n |y_i - \beta_0 - c/\sqrt{nh}|^{p-1} \mbox{sign}(y_i - \beta_0 - c/\sqrt{nh}) K_h(x_i - x) \geq 0\right)
	\nonumber  \\
	&= & P\left(\sum_{i=1}^n |y_i - \beta_0 - c/\sqrt{nh}|^{p-1} \mbox{sign}(y_i - \beta_0 - c/\sqrt{nh}) K_h(x_i - x) \leq 0\right) \nonumber\\
	&=& P\left(\sqrt{\frac{h}{n}}\sum_{i=1}^n |(\varepsilon_i + m(x_i)) - m(x) - c/\sqrt{nh}|^{p-1} {\rm sign}\left(\varepsilon_i + m(x_i) - m(x) - c/\sqrt{nh}\right) \right. \nonumber \\  && \quad \quad \left. K_h(x_i - x) \leq 0 \right)\nonumber \\
	&= & P\left(V_{0n}  + (p-1) E[|\varepsilon_1|^{p-2}] \left\{\mu_2(K) \sqrt{nh^5} \left(f'(x) m'(x) + m''(x) f(x)/2\right) - c f(x)\right\} \right. \nonumber \\  && \quad \quad \left. + o(\sqrt{nh^5}) \leq 0\right) \nonumber \\
	&=& P \left(\frac{V_{0n} + (p-1)E[|\varepsilon_1|^{p-2}] (m''(x)f(x)/2 + m'(x)f'(x))\sqrt{nh^{5}}\mu_{2}(K)}{(p-1) E[|\varepsilon_1|^{p-2}] f(x)} + o(\sqrt{nh^5}) \leq c, \right) \nonumber 	
\end{eqnarray}
\end{small}
where we are using the definition of $V_{0n}$ given in Lemma \ref{lem:clt}. According to  that lemma,  
	$ \frac{V_{0n}}{\sqrt{f(x) R(K) E[|\varepsilon_1|^{2p-2}]}}$ converges in 
	distribution to the standard normal distribution, and the entire quantity on the left
	side of the inequality displayed directly above must 
	 converge to a normal distribution with mean
	$\mu_2(K)\left(\frac{f'(x)}{f(x)}m'(x) + \frac{m''(x)}{2}\right)\sqrt{nh^5}$ and variance $\frac{R(K)E[|\varepsilon_1|^{2p-2}]}{(p-1)^2 E^2[|\varepsilon_1|^{p-2}]f(x)}$. 
	By the definition of $F_n(c)$, we see that 
	$\sqrt{nh}(\widehat{\beta}_0 - \beta_0)$ converges to this same normal distribution limit and
	the result in (i) immediately follows.  
	
\item[(ii, iii)]
	Let $F_n(x)$ denote the joint distribution function of $[\sqrt{nh}(\widehat{\beta_0} - \beta_0), 
\sqrt{nh^{3}}(\widehat{\beta_1} - \beta_1)]$.  Then for any real $c_0$ and $c_1$, we have $F_n(c_0, c_1)$
\begin{small}
\begin{eqnarray}
		 &=& P( \sqrt{nh^{1+2j}}(\widehat{\beta_j} - \beta_j) \leq c_j; j = 0,1)  \nonumber \\
		&=& P\left(\frac{1}{h^j}\sqrt{\frac{h}{n}}\sum_{i=1}^n (x_i-x)^j \left|\varepsilon_i + \frac{m''(x)}{2}(x_i-x)^2 + \frac{m'''(x)}{3!}(x_i-x)^3 - \frac{c_0}{\sqrt{nh}}- \frac{c_1(x_i-x)}{\sqrt{nh^3}}\right|^{p-1} \right.  \nonumber \\
		 &&\times \left.  \mbox{sign}\left(\varepsilon_i + \frac{m''(x)}{2}(x_i-x)^2 + \frac{m'''(x)}{3!}(x_i-x)^3 - \frac{c_0}{\sqrt{nh}}- \frac{c_1(x_i-x)}{\sqrt{nh^3}}\right) K_h(x_i - x) \ge 0 \right)	 \nonumber \\
		&=& P \left(V_j +\frac{1}{h^j}\sqrt{\frac{h}{n}} \sum_{i = 1}^n (p-1)|\varepsilon_i|^{p-2}\left(\frac{m^{(2+j)}(x)}{(2+j)!}(x_i-x)^{2+2j} -\frac{c_j(x_i-x)^{2j}}{\sqrt{nh^{1+2j}}}\right) K_h(x_i - x) \ge 0 \right) \nonumber\\ 
		&=& P\left(V_j + (p-1)E[|\varepsilon_1|^{p-2}]\left(\frac{m^{(2+j)}(x)}{(2+j)!} \mu_{2 + 2j} f(x) \sqrt{nh^{5 + 2j}}-\mu_{2j}(K)f(x)c_j\right) \leq 0; j = 0, 1\right) \nonumber \\
		&=& P \left(\frac{V_j + (p-1)E[|\varepsilon_1|^{p-2}] m^{(2+j)}(x)\sqrt{nh^{5+2j}}\mu_{2+2j}(K)f(x)}{(p-1)(2+j)! E[|\varepsilon_1|^{p-2}]\mu_{2j}(K) f(x)} \leq c_j;  j = 0, 1\right)\nonumber 
	\end{eqnarray}
\end{small}
Therefore,
$\sqrt{nh}(\widehat{\beta_0} -\beta_0)$ converges to the same limit as $\frac{V_0 + f(p-1)E[|\varepsilon_1|^{p-2}] m''(x)/2\sqrt{nh^5}\mu_2}{f(p-1)E[|\varepsilon_1|^{p-2}]}$ so $\widehat{\beta_0}$ is asymptotically normal with mean  $m(x) + \frac{h^2}{2} \mu_2(K)m''(x)$ and variance $\frac{R(K)\mu_{2p-2}(|\varepsilon|)}{f(x)(p-1)^2 \mu_{p-2}(|\varepsilon|)}$.

Also, 
$\sqrt{nh^3}(\widehat{\beta_1} - \beta_1)$ is asymptotically normal with the same limit distribution as $\frac{V_1}{(p-1)E[|\varepsilon|^{p-2}]f(x)\mu_2}+\frac{m'''(x)\sqrt{nh^7}\mu_4}{3!\mu_2}$, that is, $N\left(\frac{m'''(x)\sqrt{nh^7}\mu_4}{3!\mu_2}, \frac{\mu_2^2(K^2)E[|\varepsilon|^{2p-2}]}{f(x)\mu_2(K)(p-1)^2 \,u_{p-2}^2(|\varepsilon|)}\right)$ so $\widehat{\beta_1}$ is asymptotically normal with mean $m'(x)+m'''(x)/3!h^2\mu_4(K)/\mu_2(K)$ and variance $\frac{\mu_2^2(K^2)E[|\varepsilon|^{2p-2}]}{nh^3f(x)\mu_2(K)(p-1)^2 \mu_{p-2}^2(|\varepsilon|)}$.

Asymptotic independence of $\widehat{\beta}_0$ and $\widehat{\beta}_1$ then follows from Lemma
 \ref{lem:clt}.  
	\end{enumerate}
\end{proof}

\section{Bandwidth Selection}\label{sec:bw}
Our bandwidth selection method leverages the availability of good bandwidth choices for LLS regression ($h_2$).   Specifically, we obtain the value for $h_2$ from the method outlined by \citet{ruppert1995effective} which is  an effective approach for bandwidth selection in LLS regression. \par
 
Referring to the remarks made in Section \ref{theory}, we can establish that the relationship expressed in equation (\ref{eq:hp_eps}) remains valid for the optimal bandwidth in local polynomial $L_2$-norm and $L_p$-norm regression. Given that MSE combines the squared bias and variance of a model, we can find an MSE-optimal
bandwidth for local $L_p$-norm regression by differentiating the MSE with respect to $h$ and solving the resulting equation.   The optimal value $h_p$ satisfies
\begin{equation}
	h_p^5 = \frac{E[|\varepsilon_1|^{2p-2}]}{E[|\varepsilon_1|^{p-2}]^2 (p-1)^2 E[|\varepsilon_1|^2]} h_2^5.
	\label{eq:hp_eps}
	\end{equation}

Adapting moment properties derived by \citet{giacalone2020combined}, we can also demonstrate that the following holds  when errors follow the GED.
\begin{equation}
	h_p^5 = \frac{\Gamma(\frac{1}{p})^2 \Gamma(\frac{2p-1}{p})}{\Gamma(\frac{p-1}{p})^2\Gamma(\frac{3}{p}) (p-1)^2}h_2^5.
	\label{eq:hp_gamma}
	\end{equation}

\section{Estimating P}\label{sec:p}

\begin{figure}[h!]
	\begin{center}
		
		\includegraphics[scale=0.55]{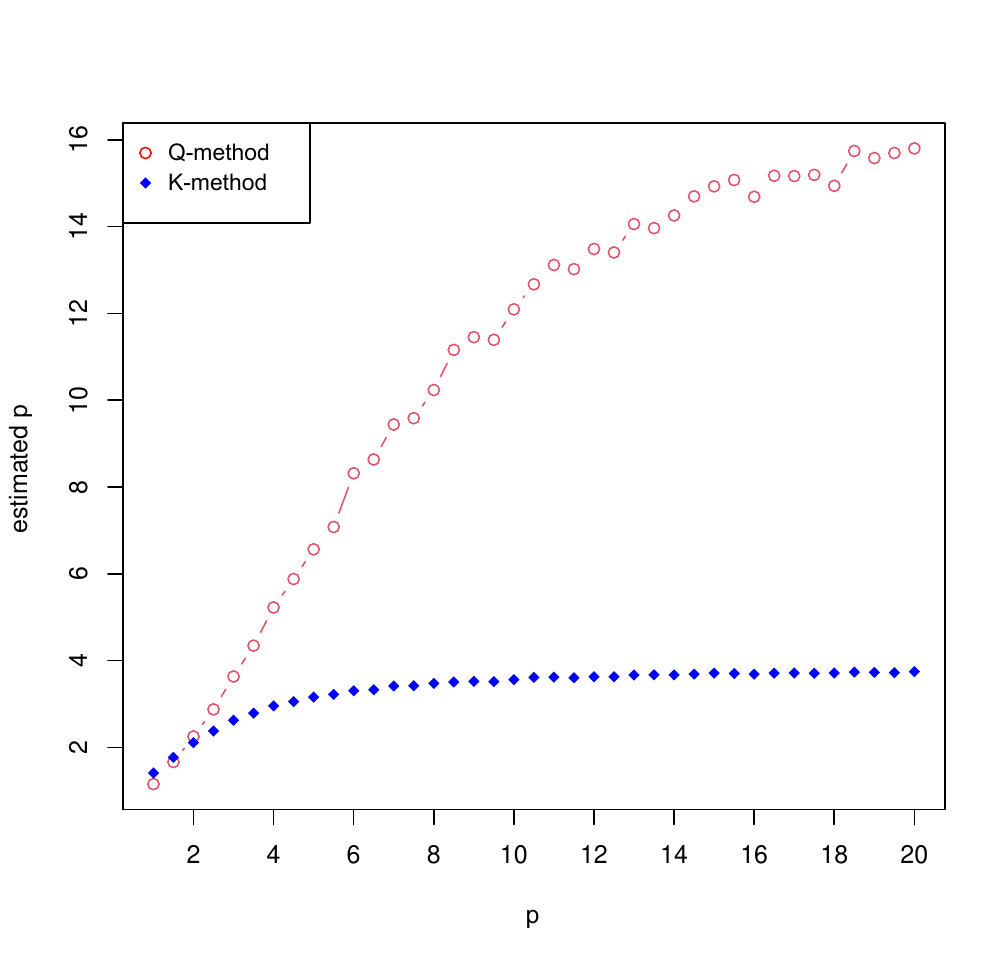}
		\caption{Comparison of the estimated $p$ values by two methods. The results are obtained by averaging 1000 simulations for sample size 100, for varying $p \in [1,20]$.}
		\label{fig:comp_p}
	\end{center} 
\end{figure}

Building upon the outcomes of a simulation study by \citet{mineo2003estimation} that highlights the most effective approach for estimating the shape parameter $p$ relies on the utilization of Generalized Kurtosis ($\kappa$), we initially employed the formula given by \citet{money1982linear} to gauge $p$ estimation. The authors of that work showed that if the kurtosis of an error distribution is known or can be estimated, then using formula (\ref{eq:p}) will yield good regression estimates for a wide variety of unknown error distributions.
\begin{equation}\label{eq:p}
	\hat{p} = \frac{9}{\kappa^2}+1 \hspace{1cm} 1\le p < \infty
\end{equation}
We refer to this method as the $K$ method. We conducted a simulation study involving a large range of $p$ values, from very heavy-tailed ($p = 1$) to thin-tailed distributions ($p = 20$). For each population,
1000 samples were generated and the corresponding $p$ values were estimated using $K$ method. We observed that as $p$ gets larger, 
 the estimated kurtosis ($\kappa$) stabilizes and fluctuates moderately at values slightly less than 2.0, leading to constrained and biased estimates of $p$, as shown in figure \ref{fig:comp_p}. This limitation is consistent with other moment-based methods, which become unreliable when the underlying error distribution deviates from their assumptions.

This leads us to consider another method for estimating $p$ which is expected to work better, especially when $p$ is supposed to be large. We will refer to this method as the $Q$ method. Our proposed $Q$ method completely removes the reliance on moments by estimating $p$ directly from residual distributions. It utilizes residuals from a fitted model coming from LLS. It determines $p$ by regressing various quantiles for candidate G.E.D, derived from a sequence of potential $p$'s, against the residuals, as expressed in the equation below:
\[q = \beta q_{p} +\varepsilon,\]
where $q$ are the quantiles of the residual, and $q_{p}$ are quantiles from GED with shape parameter $p$. The $p$ that minimizes the $L_2$-norm error, i.e.  \(\sum_{i=1}^{n}(q_i - \hat{\beta} q_{i(p)})^2\),  renders our selected value of $p$. Our method is detailed in Algorithm \ref{alg:estp} in the Appendix.\par
Using the same simulation study, we estimated $p$ by $Q$ method. The results illustrated in figure \ref{fig:comp_p} reveal the $Q$ method reduced systematic errors. These outcomes substantiate the assertion that the $Q$ method excels in comparison to $K$ method, particularly when handling larger $p$ values. 

 In summary, our fitting method includes three steps:
\begin{enumerate}
\item We initially determine the optimal bandwidth $h_p$. 
\item Employing $p=2$ and $h_p$, we calculate $\hat{p}$ from the residuals using the $Q$ method.
\item We then estimate the local $L_p$-norm regression curve by minimizing (\ref{eq:lc}) or (\ref{eq:llp}). 
\end{enumerate}

\section{Simulation Studies and Examples}\label{sec:sim}
We evaluate the performance of our proposed method, particularly local linear $L_p$ norm estimator (LLP), on both simulated and real-life data. While the choice of kernel is known to be of less importance in classical kernel regression analysis, we employ Gaussian kernel function throughout our experiments.\par
\subsection{Simulation Studies}
For one-dimensional data, we evaluate the performance LLP regression under four different error distributions: uniform, triangular, symmetric bimodal, and GED with shape parameters $p \in [1,10]$. We systematically assess LLP across these distributions by simulating 1000 replicates for each of four target functions and three sample sizes ($n = 50, 100, 200$). The design points ($x$) are randomly drawn from a uniform distribution over $[0,1]$.\par 
To compare the performance of LLS and LLP estimators in terms of Mean Squared Error (MSE). We reported MSE values represent the average over 1000 simulations. To mitigate boundary effects from skewing the results, we restrict our evaluation on the interval $[0.05, 0.95]$. The target regression functions are as follows:
\begin{enumerate}
	\item $m(x) = \sin(x)$,
	\item $ m(x) = x+ 2^{-16x^2}$,
	\item $ m(x) =\sin(2x) + 2^{-16x^2}$,
	\item $ m(x) = 0.3e^{-4(x+1)^2} + 0.7e^{-16(x-1)^2}$
\end{enumerate}

\subsection{Results}

\begin{figure}[h]
	\begin{center}
		\includegraphics[scale= 0.75]{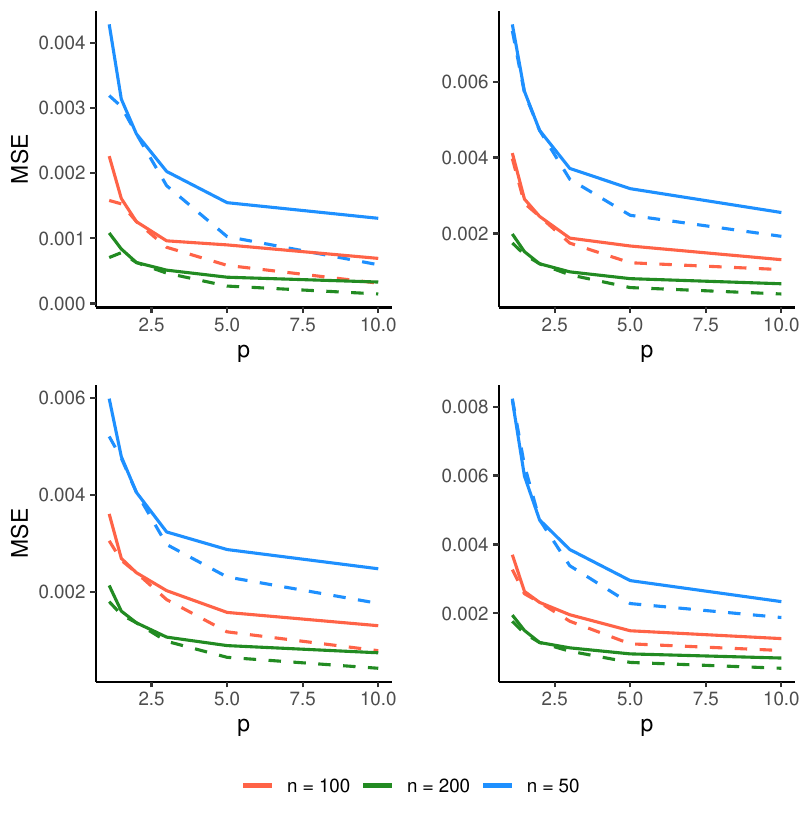} \caption{Comparison of average MSEs of LLS and LLP for GED error distribution ($\mu$=0, $\sigma_p$ = 0.2). Plots (a) to (d) correspond to function 1 to 4, respectively. Solid lines represent local least squares estimator (LLS) and dashed lines show local linear $L_p$-norm estimator (LLP).}
		\label{fig:plots}
	\end{center}
\end{figure}

Figure \ref{fig:plots} represents a comparison of MSEs of LLS and LLP as per the simulation outcomes for all four target functions. As anticipated, the performance of nonparametric regression methods generally improves with increasing sample size. Our findings indicate that our method consistently outperforms the LLS  across three distinct sample sizes, particularly for higher values of $p$. Conversely, when the estimated value of $p$ is close to 2, our method performs similarly to LLS, as expected for data distributions that closely resemble the normal distribution. Refer to the Appendix B for results of other error distributions.

In addition, the simulation results for error of uniform distributions are shown in table \ref{uni_results}. Our method shows significant improvement over LLS. We report the estimated shape parameter in column $\hat{p}$ using $Q$ method as well as the average of optimal bandwidths ($h_2, h_p$) and MSE values for LLS and LLP from 1000 simulations. We observe that $\hat{p}$ is increasing with the sample size. This is to be expected, since the optimal value of $p$ should be infinite for the uniform distribution. 

\begin{table}[h!]
	\centering
	\small
	\begin{tabular}{ccccccc}
		Function & $n$ & $\hat{p}$ & $h_2$ & $MSE_{lls}$ & $h_p$ & $MSE_{llp}$ \\
		\midrule
		 & 50  & 9.65 & 0.10 & 0.025 & 0.08 & 0.014 \\
		1 & 100 & 11.62 & 0.09 & 0.014 & 0.07 & 0.005 \\
		 & 200 & 14.16 & 0.10 & 0.007 & 0.07 & 0.002 \\
		\midrule
		 & 50  & 9.05 & 0.09 & 0.027 & 0.08 & 0.016 \\
		2 & 100 & 11.19 & 0.08 & 0.015 & 0.07 & 0.006 \\
		 & 200 & 13.73 & 0.08 & 0.007 & 0.06 & 0.002 \\
		\midrule
		 & 50  & 9.25 & 0.10 & 0.027 & 0.08 & 0.016 \\
		3 & 100 & 11.26 & 0.09 & 0.015 & 0.07 & 0.006 \\
		 & 200 & 13.88 & 0.09 & 0.007 & 0.07 & 0.002 \\
		\midrule
		 & 50  & 9.23 & 0.09 & 0.027 & 0.08 & 0.016 \\
		4 & 100 & 13.34 & 0.09 & 0.016 & 0.07 & 0.006 \\
		 & 200 & 13.87 & 0.09 & 0.008 & 0.06 & 0.002 \\
		\bottomrule
	\end{tabular} 
\caption{Simulation results for uniform error distribution ($-0.5,0.5$)}
\label{uni_results}
\end{table}

\subsection{Application to Real Life Data}
We employed two distinct datasets in our analysis to evaluate the performance of our proposed method, specifically the local linear $L_p$-norm estimator, in comparison to local least squares in real-world applications. In both cases, we compared our method against the \textit{locpoly} function from the KernSmooth package \citep{kernSmooth}, which utilizes local polynomial regression.

The first dataset, which we refer to as the Beluga dataset, represents heavy-tailed data, with the shape parameter estimated to be relatively small. Using our $Q$ method $p$ estimated to be 1.35. The dataset consists of 228 records from a study on the nursing behaviors of two Beluga whale calves born in captivity \citep{russell1997nursing}. It is a time series in which the target variable of interest is the duration of nursing (time spent latched on per period) for one of the calves. The square root transformation taken to stabilize variance.

Figure \ref{fig:both} (a) illustrates the fitted curves obtained using both local least squares and local linear $L_p$ norm estimator with its corresponding bootstrap confidence bands, described in the next subsection.  We have {excluded the estimates in the boundary regions, since our focus has been exclusively on
evaluation points in the interior.  There are subtle differences between the LLS and LLP curves, particularly in the period between Day 140 and Day 200.  The LLS curve actually falls outside the LLP confidence bands around Day 40 and again, between Days 150 and 160.} 
\subsubsection{Bootstrap Confidence Bands}

To compute pointwise bootstrap confidence bands, we employ a variant of the basic $1 - 2\alpha$
bootstrap confidence interval described by \citet{davison1997bootstrap}:
\[ (2\widehat{m}(x) - \widehat{\widehat{m}}(x)_{1-\alpha}, 2\widehat{m}(x) - \widehat{\widehat{m}}(x)_{\alpha}).  \]
Here, 
$\widehat{\widehat{m}}(x)_{\alpha}$ is the lower $\alpha$ quantile of the bootstrap distribution
of the estimator.   Because the estimator is biased, \citet{davison1997bootstrap} recommend
a modified form of bootstrap simulation.    Bootstrap-simulated responses used in the calculation
of $\widehat{\widehat{m}}(x)$ are of the form
\[ y_i^\star = \widehat{m}(x_i) + e_i^\star,  \ \ \ \ i = 1, 2, \ldots, n \]
where the bootstrap residuals $e^\star$ are resampled from residuals calculated for a bias-reduced
estimate of $m(x)$.  The method of \citet{cheng2018bias} is simple and effective.    Furthermore,
based on the theoretical results obtained earlier in this paper, the method applies to the local Lp-norm
estimator.    Essentially, the method employed here is to fit the following simple regression model
\[ \widehat{m}(x; h) = \beta_0 + \beta_1 h^2 + \epsilon \]
where the ``data'' are a collection of local Lp-norm estimates of the regression function 
corresponding to a set  of bandwidths $h$.  The range of bandwidths is $(h/2, 2h)$ where
$h$ is the asymptotically optimal bandwidth.  Because of the asymptotic form of the bias
in $\widehat{m}(x)$, the least-squares estimate of $\beta_0$ serves as an estimator of $m(x)$
which has bias of $o(h^2)$ instead of $O(h^2)$.  \par

Secondly, we incorporated a distinct dataset into our analysis, which represents an example of light-tailed data, with the parameter $p$ estimated to be relatively large. The dataset consists of measurements of four variables: height (mm), width (mm), diagonal (cm), and an index variable for 51 randomly selected Mathematics and Statistics books. The index variable is computed as the sum of the squares of the height and width for each book.

We conducted a regression analysis between the diagonal and the square root of the index, using the latter as the predictor variable while assuming a uniform error distribution over $[-0.5, 0.5]$. Using our proposed Q method, the parameter $p$ was estimated to be approximately 19.5.

\begin{figure}[h]
	\begin{center}
		\includegraphics[scale=1]{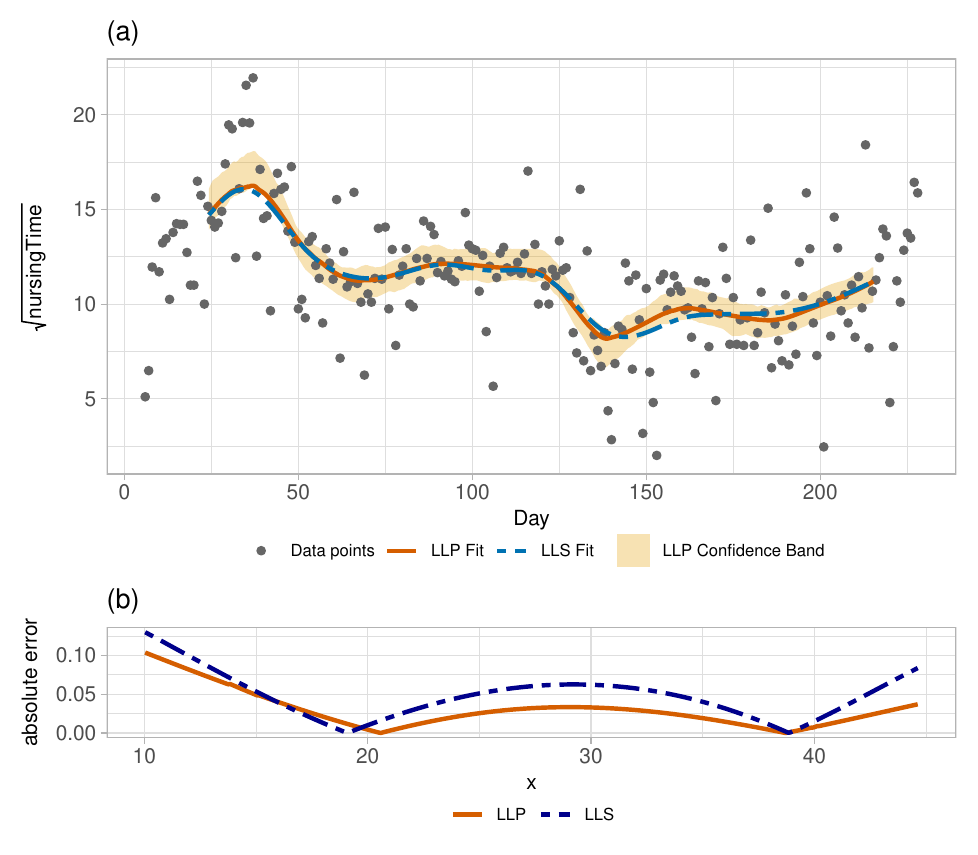}
		\caption{(a) Comparison of fitted line by Local Least Squares and Local Linear $L_p$ norm estimator with the confidence band for Beluga dataset. (b) Comparison of absolute error between Local Linear $L_p$ norm estimator (solid line) and Local Least Squares (dotted line) for Rectangles dataset.}
		\label{fig:both}
	\end{center} 	
\end{figure}

A closer examination of the absolute error between the two methods is presented in figure \ref{fig:both}(b). Our method consistently produces lower errors than than $locpoly$ across most of the input domain. According to our theoretical framework, bias is linked to the second derivative of the regression function. Specifically, when $m(x) = \sqrt{x}$ the second derivative is given by:
\[m''(x) = -\frac{1}{x\sqrt{x}}\]

Interestingly, our results confirm this relationship: the discrepancy is more pronounced at the lower end of the input range, where the second derivative is larger, indicating greater bias. Conversely, at the upper end, we observe a notable reduction in bias, aligning with our theoretical expectations. This finding reinforces the validity of our approach and further demonstrates its effectiveness in reducing estimation error.

\section{Conclusion and Future Directions}\label{sec:conc}

In many real-life applications, such as economics, finance, and health sciences, regression analysis is an "off-the-shelf" solution. However, regression models produce more accurate estimates only if certain assumptions are met. A significant role is played by the shape of the error distribution in obtaining efficient estimation of the actual relationship in the data set. The error distribution is typically approximated by Gaussian, which is not the case in many applications. In addition, parametric regression assumes yet another assumption about the form of the regression function, which may not always hold true. This research relaxes these two assumptions and introduces a more flexible method by using the $L_p$-norm estimator in the context of local polynomial regression.\par
We found that local $L_p$-norm estimator outperformed local least squares regression when there was a non-Gaussian error. In other words, when the distribution of data is either Platykurtic  or Leptokurtic, local polynomial $L_p$-norm regression works better than local least squares. Based on our theoretical results which show that the $L_p$-norm method can yield asymptotically normal estimates with the asymptotic bias as local least squares, we derived a quick method for selecting an appropriate bandwidth $h_p$.  Additionally, we showed that the sample moment-based method developed by \citet{money1982linear} for estimating $p$ is not consistently robust across a range of values of $p$. Our simulation analysis demonstrated that our proposed $Q$ method outperforms it.  \par 

To assess the adaptability of our local polynomial $l_p$-norm regression method in higher dimensions, we conducted a preliminary study on two-dimensional data using the Generalized Error Distribution under varying error conditions. Our results indicate that LLP outperforms LLS, particularly for larger values of $p$. A summary of the simulation results, including MSE comparisons, is provided in Appendix B (table \ref{n200_2d}), demonstrating the promising performance of our method in 2D. Further exploration in this direction is needed.

\bibliography{mybibfile}

\appendix
\newpage
\section{Algorithms} \label{App:algs}

We employed the \textit{nlm()} function from the \textit{stats} package \citep{stats}. In cases where the derivative of the function is desired,  we can modify the algorithm to return \textit{estimate[2]} which yields the first derivative of the function, represented by $\beta_1$. \\
\begin{algorithm}[h] 
	\scriptsize
	\begin{algorithmic}  \label{alg:loclp}
		\Require {$x, y$ : sample data , 	$h$:  bandwidth , $xpoints$ : equally spaced data points, $lp$ : power,
			$order$ : Local constant(0) or local linear (1) }
		\Ensure { xpoints, estimated ypoints }
		\Function{Loc\_lp}{$x, y, h, xpoints ,lp, order$}
		\newline
		\State $ypoints$ : An array of length of xpoints
		\State $starting\_value = 0$
		\State $obj\_func1 \gets |y -\beta_0|^p K_h(x - xpoint)$
		\State $obj\_func2 \gets |y -\beta_0 -\beta_1(x - xpoint)|^p K_h(x - xpoint)$
		\newline
		\For{$i \in 1 : length(xpoints)$}
		\State $ xpoint = xpoint[i]$
		\If{$order$ == 0}
		\State $starting\_value = mean(y[|x-xpoint|<2*h])$
		\State $ypoints[i] = nlm(obj\_func1,start\_value)\$estimate[1]$
		\ElsIf {$order$ == 1}
		\State $xx = x- xpoint$
		\State $b = lm( y ~ xx, weights = dnorm(xx, sd = h) )$
		\State $starting\_value = coef(b)$
		\State $ypoints[i] = nlm(obj\_func2, start\_value)\$estimate[1]$
		\EndIf
		\EndFor
		\State \Return { $xpoints, ypoints$}
		\EndFunction
	\end{algorithmic}
	\caption{Local Polynomial $L_p$ Norm Regression}
	\label{alg:llp}
\end{algorithm}

\begin{algorithm}
	\caption{Estimating P }\label{alg:estp}
	\scriptsize
	\begin{algorithmic}
		\Require{$resid$ :residuals of the fitted model by least-squares }
		\Ensure{$estimated$ $p$}
		\Statex
		\Function{EstimateP}{$resid$}
		\State $p$:  A list of potential p in range of [1,20]
		\State $n$ : size of $resid$ 
		\State $y.q$ : sorted residuals
		\State $e_{norm}$ : an empty numeric vector of length $p$ 
		
		\For{$i \in 1 : length(p)$}
		\State  $ normp.q = qnormp(ppoints(n), p = p[i]) $ 
		\State $ normp.QR = qr(normp.q)$
		\State $ Q = qr.Q(normp.QR, complete = TRUE)$
		\State $ eps = t(Q[,-1])\%*\% y.q $
		\State $e_{norm}[i]= norm(eps)$
		\EndFor
		\State \Return { $p[i]$ which gives the minimum $e_{norm}$}
		\EndFunction
	\end{algorithmic}
\end{algorithm}

\begin{algorithm}	
	\scriptsize
	\begin{algorithmic}
		\Require{$res$ : residuals from fitted LLS , $h_2$:  LLS bandwidth , $p$ : estimated shape parameter}
		\Ensure{$optimum$ $bandwidth$}
		\Statex
		\Function{FindBW}{$res, h_2, p$}
		\State $h_p = 0$
		\State $h_p= \left(\frac{mean(|res|)^{2p-2}}{(p-1)^2 mean(|res|^{p-2})^2 mean(res^2)}\right)^{1/5}\times h_2$
		
		\State \Return { $h_p$}
		\EndFunction
	\end{algorithmic}
	\caption{Bandwidth Selection Algorithm}
	\label{alg:bw}
\end{algorithm}

\newpage

\section{Tables} \label{App:tbls}


\subsection{Triangular Dist. ($-0.5,0.5$)}
The triangular error distribution is generated by a summation of two uniform distributions over the same range.
\begin{table}[H]
	\centering
	\caption{Target function (1), Triangular Dist. } 
	\label{case1_tri}
	\small
	\begin{tabular}{rrrrrr}
	
		$n$ & $\hat{p}$ & $h_2$ & $MSE_{lls}$ & $h_p$ & $MSE_{llp}$\\
		\hline
		50 & 3.12 & 0.10 & 0.013& 0.09 & 0.012\\
		100 & 2.98 & 0.09 & 0.007&  0.09 & 0.006\\
		200 & 2.91 & 0.09 & 0.004 & 0.09 & 0.003\\ 
		\hline
	\end{tabular}
\end{table}

\begin{table}[H]
	\centering
	\caption{Target function (2),Triangular Dist. } 
	\label{case2_tri}
	\small
	\begin{tabular}{rrrrrr}
	
		$n$ & $\hat{p}$ & $h_2$ & $MSE_{lls}$ & $h_p$ & $MSE_{llp}$\\
		\hline
		50 & 3.12 & 0.10 & 0.013& 0.09 & 0.012\\
		100 & 3.04 & 0.09 & 0.008&  0.09 & 0.006\\
		200 & 2.98 & 0.08 & 0.003 & 0.07 & 0.003\\ 
		\hline
	\end{tabular}
\end{table}

\begin{table}[H]
	\centering
	\caption{Target function (3),Triangular Dist. } 
	\label{case3_tri}
	\small
	\begin{tabular}{rrrrrr}
	
		$n$ & $\hat{p}$ & $h_2$ & $MSE_{lls}$ & $h_p$ & $MSE_{llp}$\\
		\hline
		50 & 3.05 & 0.09 & 0.014& 0.09 & 0.013\\
		100 & 2.95 & 0.09 & 0.008&  0.08 & 0.007\\
		200 & 2.90 & 0.08 & 0.004 & 0.08 & 0.003\\ 
		\hline
	\end{tabular}
\end{table}

\begin{table}[H]
	\centering
	\caption{Target function (4),Triangular Dist. } 
	\label{case4_tri}
	\small
	\begin{tabular}{rrrrrr}
		
		$n$ & $\hat{p}$ & $h_2$ & $MSE_{lls}$ & $h_p$ & $MSE_{llp}$\\
		\hline
		50 & 3.11 & 0.09 & 0.014& 0.09 & 0.013\\
		100 & 2.96 & 0.08 & 0.008&  0.08 & 0.007\\
		200 & 2.90 & 0.08 & 0.004 & 0.08 & 0.004\\ 
		\hline
	\end{tabular}
	\end{table}

\subsection{Bimodal Dist.}
We generate the bimodal error distribution by following the formula below:
\begin{equation*}
	\varepsilon = |Z| ^{\frac{1}{\alpha(1-B)}}
\end{equation*}
where $Z$ is standard normal random variable and $B$ has a Bernoulli distribution. $\alpha$ is
obtained from the sequence of $[0:4;\dots ,1:8]$.

\begin{table}[H]
	\centering
	\caption{Target function (1), Bimodal Dist. , n = 50 } 
	\label{case1_bi_50}

	\begin{tabular}{rrrrrr}
		\hline
		$\alpha$ & $\hat{p}$ & $h_2$ & $MSE_{lls}$ & $h_p$ & $MSE_{llp}$\\
		\hline
		0.4 & 1.10 & 0.10 & 0.480& 0.10 & 0.144\\
		0.6 & 1.17 & 0.10 & 0.153&  0.11 & 0.136\\
		0.8 & 1.54 & 0.10 & 0.095 & 0.11 & 0.086\\ 
		1.0 & 2.37 & 0.10 & 0.077 & 0.10 & 0.082\\ 
		1.2 & 3.6 & 0.10 & 0.068 & 0.09 & 0.064\\ 
		1.4 &5.67 & 0.10 & 0.069 & 0.09 & 0.055\\ 
		1.6 &8.69 & 0.10 & 0.061 & 0.09 & 0.040\\
		1.8 &10.92 & 0.10 & 0.064 & 0.08 & 0.033\\ 
		\hline
	\end{tabular}
\end{table}

\begin{table}[H]
	\centering
	\caption{Target function (1), Bimodal Dist., n = 100 } 
	\label{case1_bi_100}
	\small
	\begin{tabular}{rrrrrr}
		
		$\alpha$ & $\hat{p}$ & $h_2$ & $MSE_{lls}$ & $h_p$ & $MSE_{llp}$\\
		\hline
		0.4 & 1.10 & 0.08 & 0.433& 0.08 & 0.201\\
		0.6 & 1.12 & 0.09 & 0.080&  0.10 & 0.035\\
		0.8 & 1.42 & 0.09 & 0.055 &0.09 & 0.045\\ 
		1.0 & 2.17 & 0.10 & 0.045 & 0.10 & 0.047\\ 
		1.2 & 3.35 & 0.09 & 0.040 & 0.09 & 0.037\\ 
		1.4 &5.59 & 0.09 & 0.036 & 0.08 & 0.027\\ 
		1.6 &8.69 & 0.09 & 0.037 & 0.08 & 0.023\\
		1.8 &12.24 & 0.09 & 0.035 & 0.08 & 0.017\\ 
		\hline
	\end{tabular}
\end{table}

\begin{table}[H]
	\centering
	\caption{Target function (1),, Bimodal Dist., n = 200 } 
	\label{case1_bi_200}
	\small
	\begin{tabular}{rrrrrr}
	
		$\alpha$ & $\hat{p}$ & $h_2$ & $MSE_{lls}$ & $h_p$ & $MSE_{llp}$\\
		\hline
		0.4 & 1.10 & 0.09 & 0.147& 0.08 & 0.011\\
		0.6 & 1.11 & 0.09 & 0.044&  0.10 & 0.016\\
		0.8 & 1.34 & 0.09 & 0.027 &0.09 & 0.021\\ 
		1.0 & 2.10 & 0.10 & 0.021 & 0.09 & 0.021\\ 
		1.2 & 3.33 & 0.09 & 0.020 & 0.09 & 0.017\\ 
		1.4 &5.28 & 0.10 & 0.017 & 0.09 & 0.013\\ 
		1.6 &8.52 & 0.10 & 0.016 & 0.09 & 0.011\\
		1.8 &12.93 & 0.09 & 0.016 & 0.09 & 0.009\\ 
		\hline
	\end{tabular}
\end{table}
\begin{table}[H]
	\centering
	\caption{Target function (2), Bimodal Dist., n = 50 } 
	\label{case2_bi_50}
	\small
	\begin{tabular}{rrrrrr}
	
		$\alpha$ & $\hat{p}$ & $h_2$ & $MSE_{lls}$ & $h_p$ & $MSE_{llp}$\\
		\hline
		0.4 & 1.11 & 0.10 & 0.335& 0.11 & 0.053\\
		0.6 & 1.12 & 0.08 & 0.166&  0.08 & 0.099\\
		0.8 & 1.70 & 0.09 & 0.076 & 0.09 & 0.067\\ 
		1.0 & 2.68 & 0.11 & 0.076 & 0.11 & 0.074\\ 
		1.2 & 4.02 & 0.10 & 0.075 & 0.10 & 0.069\\ 
		1.4 &6.63 & 0.10 & 0.067 & 0.09 & 0.057\\ 
		1.6 &8.83 & 0.09 & 0.056 & 0.07 & 0.028\\
		1.8 &11.43 & 0.10 & 0.078 & 0.09 & 0.035\\ 
		\hline
	\end{tabular}
\end{table}

\begin{table}[H]
	\centering
	\caption{Target function (2), Bimodal Dist., n = 100 } 
	\label{case2_bi_100}
	\small
	\begin{tabular}{rrrrrr}
	
		$\alpha$ & $\hat{p}$ & $h_2$ & $MSE_{lls}$ & $h_p$ & $MSE_{llp}$\\
		\hline
		0.4 & 1.10 & 0.10 & 0.245& 0.11 & 0.043\\
		0.6 & 1.12 & 0.08 & 0.126&  0.08 & 0.098\\
		0.8 & 1.85 & 0.08 & 0.086 & 0.08 & 0.077\\ 
		1.0 & 2.68 & 0.10 & 0.076 & 0.10 & 0.075\\ 
		1.2 & 4.31 & 0.10 & 0.069 & 0.10 & 0.057\\ 
		1.4 &6.44 & 0.09 & 0.066 & 0.08 & 0.056\\ 
		1.6 &8.36 & 0.09 & 0.080 & 0.07 & 0.039\\
		1.8 &12.06 & 0.10 & 0.084 & 0.09 & 0.043\\ 
		\hline
	\end{tabular}
\end{table}

\begin{table}[H]
	\centering
	\caption{Target function (2), Bimodal Dist., n = 200 } 
	\label{case2_bi_200}
	\small
	\begin{tabular}{rrrrrr}
	
		$\alpha$ & $\hat{p}$ & $h_2$ & $MSE_{lls}$ & $h_p$ & $MSE_{llp}$\\
		\hline
		0.4 & 1.12 & 0.10 & 0.178& 0.11 & 0.099\\
		0.6 & 1.14 & 0.08 & 0.166&  0.08 & 0.098\\
		0.8 & 1.91 & 0.08 & 0.082 & 0.08 & 0.082\\ 
		1.0 & 2.59 & 0.10 & 0.083 & 0.10 & 0.082\\ 
		1.2 & 4.56 & 0.10 & 0.082 & 0.10 & 0.051\\ 
		1.4 &6.37 & 0.10 & 0.080 & 0.08 & 0.056\\ 
		1.6 &8.80 & 0.09 & 0.068 & 0.07 & 0.029\\
		1.8 &12.87 & 0.10 & 0.056 & 0.09 & 0.021\\ 
		\hline
	\end{tabular}
\end{table}

\begin{table}[H]
	\centering
	\caption{Target function (3), Bimodal Dist., n = 50 } 
	\label{case3_bi_50}
	\small
	\begin{tabular}{rrrrrr}
	
		$\alpha$ & $\hat{p}$ & $h_2$ & $MSE_{lls}$ & $h_p$ & $MSE_{llp}$\\
		\hline
		0.4 & 1.10 & 0.10 & 0.481& 0.11 & 0.353\\
		0.6 & 1.17 & 0.10 & 0.154&  0.11 & 0.102\\
		0.8 & 1.54 & 0.09 & 0.096 & 0.09 & 0.087\\ 
		1.0 & 2.35 & 0.11 & 0.077 & 0.11 & 0.082\\ 
		1.2 & 3.60 & 0.10 & 0.069 & 0.09 & 0.067\\ 
		1.4 &5.63 & 0.09 & 0.070 & 0.09 & 0.055\\ 
		1.6 &8.57 & 0.10 & 0.062 & 0.09 & 0.041\\
		1.8 &10.99 & 0.10 & 0.064 & 0.08 & 0.035\\ 
		\hline
	\end{tabular}
\end{table}

\begin{table}[H]
	\centering
	\caption{Target function (3), Bimodal Dist., n = 100 } 
	\label{case3_bi_100}
	\small
	\begin{tabular}{rrrrrr}
	
		$\alpha$ & $\hat{p}$ & $h_2$ & $MSE_{lls}$ & $h_p$ & $MSE_{llp}$\\
		\hline
		0.4 & 1.10 & 0.08 & 0.433& 0.08 & 0.179\\
		0.6 & 1.12 & 0.09 & 0.154&  0.10 & 0.102\\
		0.8 & 1.42 & 0.09 & 0.055 & 0.09 & 0.046\\ 
		1.0 & 2.17& 0.09 & 0.046 & 0.09 & 0.048\\ 
		1.2 & 3.34 & 0.09 & 0.041 & 0.09 & 0.038\\ 
		1.4 &5.89 & 0.09 & 0.037 & 0.09 & 0.028\\ 
		1.6 &8.63 & 0.09 & 0.038 & 0.08 & 0.023\\
		1.8 &12.40 & 0.09 & 0.035 & 0.07 & 0.017\\ 
		\hline
	\end{tabular}
\end{table}

\begin{table}[H]
	\centering
	\caption{Target function (3), Bimodal Dist., n = 200 } 
	\label{case3_bi_200}
	\small
	\begin{tabular}{rrrrrr}
		
		$\alpha$ & $\hat{p}$ & $h_2$ & $MSE_{lls}$ & $h_p$ & $MSE_{llp}$\\
		\hline
		0.4 & 1.10 & 0.09 & 0.147& 0.08 & 0.010\\
		0.6 & 1.10 & 0.09 & 0.047&  0.09 & 0.012\\
		0.8 & 1.34 & 0.09 & 0.028 & 0.09 & 0.021\\ 
		1.0 & 2.09& 0.09 & 0.021 & 0.09 & 0.022\\ 
		1.2 & 3.32 & 0.09 & 0.020 & 0.09 & 0.018\\ 
		1.4 &5.27 & 0.09 & 0.018 & 0.09 & 0.014\\ 
		1.6 &8.48 & 0.09 & 0.017 & 0.08 & 0.011\\
		1.8 &12.93 & 0.09 & 0.017& 0.08 & 0.009\\ 
		\hline
	\end{tabular}
\end{table}
\begin{table}[H]
	\centering
	\caption{Target function (4), Bimodal Dist., n = 50} 
	\label{case4_bi_50}
	\small
	\begin{tabular}{rrrrrr}
		
		$\alpha$ & $\hat{p}$ & $h_2$ & $MSE_{lls}$ & $h_p$ & $MSE_{llp}$\\
		\hline
		0.4 & 1.10 & 0.10 & 0.485& 0.11 & 0.111\\
		0.6 & 1.16 & 0.09 & 0.154&  0.12 & 0.098\\
		0.8 & 1.58 & 0.10 & 0.090 & 0.10 & 0.082\\ 
		1.0 & 2.51 & 0.09 & 0.073 & 0.09 & 0..079\\ 
		1.2 & 3.60 & 0.08 & 0.067 & 0.08 & 0.062\\ 
		1.4 &5.29 & 0.10 & 0.077 & 0.08 & 0.053\\ 
		1.6 &8.90& 0.10& 0.059 & 0.08 & 0.043\\
		1.8 &11.57 & 0.09 & 0.060 & 0.08 & 0.035\\ 
		\hline
	\end{tabular}
\end{table}
\begin{table}[H]
	\centering
	\caption{Target function (4), Bimodal Dist., n = 100} 
	\label{case4_bi_100}
	\small
	\begin{tabular}{rrrrrr}
		
		$\alpha$ & $\hat{p}$ & $h_2$ & $MSE_{lls}$ & $h_p$ & $MSE_{llp}$\\
		\hline
		0.4 & 1.11 & 0.09 & 0.211& 0.09 & 0.019\\
		0.6 & 1.12 & 0.09 & 0.056&  0.10 & 0.022\\
		0.8 & 1.34 & 0.09 & 0.049 & 0.09 & 0.032\\ 
		1.0 & 2.08 & 0.09 & 0.041 & 0.09 & 0..041\\ 
		1.2 & 3.47 & 0.08 & 0.038 & 0.08 & 0.020\\ 
		1.4 &5.71 & 0.09 & 0.037 & 0.09 & 0.020\\ 
		1.6 &8.69 & 0.09& 0.036 & 0.09 & 0.021\\
		1.8 &12.84 & 0.09 & 0.027 & 0.08 & 0.008\\ 
		\hline
	\end{tabular}
\end{table}

\begin{table}[H]
	\centering
	\caption{Target function (4), Bimodal Dist., n = 200} 
	\label{case4_bi_200}
	\small
	\begin{tabular}{rrrrrr}
		
		$\alpha$ & $\hat{p}$ & $h_2$ & $MSE_{lls}$ & $h_p$ & $MSE_{llp}$\\
		\hline
		0.4 & 1.10 & 0.09 & 0.147& 0.08 & 0.010\\
		0.6 & 1.17 & 0.09 & 0.045&  0.09 & 0.012\\
		0.8 & 1.34 & 0.09 & 0.028 & 0.09 & 0.021\\ 
		1.0 & 2.10 & 0.09 & 0.022 & 0.09 & 0.022\\ 
		1.2 & 3.31 & 0.09 & 0.020 & 0.08 & 0.018\\ 
		1.4 &5.24 & 0.09 & 0.018 & 0.09 & 0.014\\ 
		1.6 &8.49 & 0.09& 0.017 & 0.09 & 0.011\\
		1.8 &12.84 & 0.09 & 0.017 & 0.08 & 0.009\\ 
		\hline
	\end{tabular}
\end{table}

\subsection{2D data}

In a preliminary investigation of two-dimensional data, we focused on a single target function, employing the Generalized Error Distribution with a few selected shape parameters to represent different error conditions. This approach allowed us to assess the adaptability and performance of our proposed local polynomial $L_p$-norm regression method to higher-dimensional data under varying error scenarios. The design points $(x_1, x_2)$ for the 2D simulation were generated from a uniform distribution over \([0,1] \times [0,1]\). The target function for the 2D data simulations was:
\begin{enumerate}
	\item $m(x_1, x_2) = \sin(3x_1x_2)$ 
\end{enumerate} 

According to the simulation studies, our method shows better performance compared to LLS particularly for larger values of $p$. A summary of these simulations is presented in the Table (\ref{n200_2d}) which includes the MSE's and standard deviation of the corresponding MSE's for LLS and our method (LLP). This stimulation study shows promising outcomes of the adaptability of our proposed method in higher dimensions. Further research in this direction is needed.

\begin{table}[H]
	\centering
	\begin{tabular}{rrrrrr}
		
		$p$  & $MSE_{lls}$& $Std_{MSE_{lls}}$ & $MSE_{llp}$  & $Std_{MSE_{llp}}$ \\
		\hline
		1.1  & 19.31 & 8.89 & 15.49 & 7.62 \\
		1.5  & 13.60 & 6.05 & 12.87 & 5.92 \\
		2.0  & 10.70 & 4.62 & 10.70 & 4.62 \\
		3.0  & 8.18 & 3.67 & 7.33 & 3.34 \\
		5.0  & 6.51 & 2.87 & 4.15 & 1.90 \\
		10.0 & 5.19 & 2.27 & 1.93 & 0.87 \\
		\hline
	\end{tabular}
	\caption{Two-dimensional data with GED ($mu=0$, $\sigma_p = 0.2 $. MSE values are the average of 1000 simulation. Both MSE and Standard Deviation (Std) values presented in the table are multiplied by $10^4$. }
	\label{n200_2d}
\end{table}  

\end{document}